\documentclass[11pt,a4paper]{article}
\usepackage[nohyperref]{acl2017}
\usepackage{times}
\usepackage{latexsym}
\usepackage{mathtools}
\usepackage{multirow}
\usepackage{xspace}
\usepackage{amsmath,amssymb}
\usepackage{url}

\aclfinalcopy % Uncomment this line for the final submission
\newcommand{\vect}[1]{\mbox{\boldmath $#1$}}
\newcommand{\vocab}{\mathcal{W}}
\newcommand{\reals}{\mathbb{R}}
\newcommand{\sent}{\vect{x}}
\newcommand{\senti}{\sent^{(i)}}
\newcommand{\qtok}{Q&}
\newcommand{\ntok}[1]{#1&}
\newcommand{\enc}{f}

\newcommand{\encargs}{(\sent, j)}
\newcommand{\encfull}{\enc\encargs}
\newcommand{\encseq}{\enc_{\mathrm{FF}}}
\newcommand{\encseqfull}{\encseq\encargs}

\newcommand{\dec}{g}
\newcommand{\loss}{\mathrm{loss}}

\newcommand{\losswre}{\loss_{\mathrm{WRE}}}
\newcommand{\distance}{\Delta}
\newcommand{\typedim}{d}
\newcommand{\tedim}{d'}
\newcommand{\blwindow}{w}
\newcommand{\tewindow}{w'}

\newcommand{\octtrain}{\textsc{Oct27Train}\xspace}
\newcommand{\octdev}{\textsc{Oct27Dev}\xspace}
\newcommand{\octtest}{\textsc{Oct27Test}\xspace}
\newcommand{\dailytest}{\textsc{Daily547}\xspace}
\newcommand{\testnew}{\textsc{Test-New}\xspace}

\title{%Token Embeddings: \\ 
Learning to Embed Words in Context for  Syntactic Tasks}%Natural Language Processing}

\author{Lifu Tu \ \ \ \ \ \  Kevin Gimpel \ \ \ \ \ \ Karen Livescu\\
[1ex]
Toyota Technological Institute at Chicago, Chicago, IL, 60637, USA\\
[1ex]
{\small  
{\tt \{lifu,kgimpel,klivescu\}@ttic.edu}
}\\}

\date{}

\usepackage[english]{babel}
\usepackage{color}
\usepackage{graphicx}
\definecolor{mygreen}{RGB}{34,189,64}

\newcommand{\argmax}{\operatornamewithlimits{argmax}}

\begin{document}
\maketitle
\begin{abstract}
  We present models for embedding words in the context of surrounding words.  Such models, which we refer to as \textbf{token embeddings}, represent the characteristics of a word that are specific to a given context, such as word sense, syntactic category, and semantic role. 
We explore simple, efficient token embedding models based on standard neural network architectures.  
We learn token embeddings on a large amount of unannotated text and evaluate them as features for part-of-speech taggers and dependency parsers trained on much smaller amounts of annotated data.  We find that predictors endowed with token embeddings consistently outperform baseline predictors across a range of context window and training set sizes.
\end{abstract}

\section{Introduction}

%Vector representations of words, also called 
Word embeddings have enjoyed a surge of popularity in natural language processing (NLP) 
%for decades, and have enjoyed a surge of popularity 
due to the effectiveness of deep learning and the availability of pretrained, downloadable models for embedding words. 
Many embedding models have been 
%Researchers have developed many embedding models
developed~\cite{collobert2011natural,NIPS2013_5021,pennington-socher-manning:2014:EMNLP2014} and 
%showed how they can 
have been shown to 
improve performance on NLP tasks, including part-of-speech (POS) tagging, named entity recognition, semantic role labeling, dependency parsing, and machine translation~\cite{turian2010word,collobert2011natural,bansal-gimpel-livescu:2014:P14-2,zou-EtAl:2013:EMNLP}.

The majority of this work has focused on a single embedding for each word type in a vocabulary.\footnote{A word type is an entry in a vocabulary, while a word token is an instance of a word type in a corpus.} We will refer to these as \textbf{type embeddings}. 
However, the same word type can exhibit a range of linguistic behaviors in different contexts. To address this, some researchers 
learn \emph{multiple} embeddings for certain word types, where each embedding corresponds to a distinct sense of the type~\cite{reisinger-mooney:2010:NAACLHLT,huang-EtAl:2012:ACL20122,tian-EtAl:2014:Coling}. 
But token-level linguistic phenomena go beyond word sense, and these approaches are only reliable for frequent words. 

Several kinds of token-level phenomena relate directly to NLP tasks. Word sense disambiguation relies on context to determine which sense is intended. 
POS tagging, dependency parsing, and semantic role labeling identify syntactic categories and semantic roles for each token. Sentiment analysis and related tasks like opinion mining seek to understand word connotations in context.

In this paper, we develop and evaluate models for embedding word tokens. Our \textbf{token embeddings} capture linguistic characteristics expressed in the context of a token. 
Unlike type embeddings, it is infeasible to precompute and store all possible (or even a significant fraction of) token embeddings.  Instead, our token embedding models are parametric, so they can be applied on the fly 
to embed any word in its context.

We focus on simple and efficient token embedding models based on local context and standard neural network architectures. We evaluate our models by using them to provide features for downstream low-resource syntactic tasks: Twitter POS tagging and dependency parsing. We show that token embeddings can improve the performance of a non-structured POS tagger to match the state of the art Twitter POS tagger of \newcite{owoputi-EtAl:2013:NAACL-HLT}. We add our token embeddings to  Tweeboparser~\cite{kong2014dependency}, improving its performance and establishing a new state of the art for Twitter dependency parsing.

\section{Related Work}
%While others have developed models to embed tokens, they are limited in several ways. 
%They mostly cluster tokens of polysemous word types, then learn a separate type vector for each cluster. 
%Therefore, they are more accurately thought of as ``multi-type embeddings'' than token embeddings.  
%They are only able to learn meaningful vectors for clusters of tokens that are sufficiently frequent in the data.  
%Typical evaluations focus on frequent polysemous words, so it is unknown how well multi-type embeddings can be learned for infrequent words or infrequent senses.  
%They also require either prespecifying the number of clusters for each word type or using nonparametric techniques to induce the number of clusters automatically.  

%We now summarize prior research in embedding words in context. 
The most common way to obtain context-sensitive embeddings is to learn separate embeddings for distinct senses of each type. 
%rather than a single one, where the entries in a set correspond to distinct senses of the word type. 
Most of these methods cluster tokens into senses and learn vectors for each cluster~\cite{vu-parker:2016:N16-1,reisinger-mooney:2010:NAACLHLT,huang-EtAl:2012:ACL20122,tian-EtAl:2014:Coling,chen-liu-sun:2014:EMNLP2014,pina-johansson-2015,wu2015sense}. 
Some use bilingual information~\cite{guo-EtAl:2014:Coling,Suster16,gonen2016semi}, nonparametric methods to avoid specifying the number of clusters~\cite{neelakantan-EtAl:2014:EMNLP2014,li-jurafsky:2015:EMNLP}, topic models~\cite{liu2015topical}, grounding to WordNet~\cite{jauhar-dyer-hovy:2015:NAACL-HLT}, or senses defined as sets of POS tags for each type~\cite{qiu2014learning}. 

%We refer to these as ``multi-sense'' type embeddings 
%or ``multi-type'' embeddings 
%to distinguish them from the token embeddings that we seek to learn in this paper. 
These ``multi-type'' embeddings are restricted to modeling phenomena expressed by a single clustering of tokens for each type. 
In contrast, token embeddings are capable of 
%beyond a single set of embeddings per type and 
modeling information that cuts across phenomena categories. 
Further, as the number of clusters grows, learning multi-type embeddings becomes more difficult due to data fragmentation. % of the data. 
%; when learning more embeddings, there is less data for learning each one. 
%Rare multi-sense words may not be seen enough to learn embeddings. 
%In our approach, 
Instead, we learn \emph{parametric} models that transform a type embedding and those of its context words into a representation for the token. 
While multi-type embeddings require more data for training, 
parametric %token embedding 
models require less.

There is prior work in developing representations for tokens in the context of unsupervised or supervised training, whether with long short-term memory (LSTM) networks~\cite{kaageback2015neural,ling-EtAl:2015:EMNLP2,choi2016,melamud2016context2vec}, convolutional networks~\cite{collobert2011natural}, or other architectures. 
%, then perform supervised classification on those tokens. For example, \newcite{ling-EtAl:2015:EMNLP2} used a bidirectional LSTM to embed tokens while training a POS tagger. 
%\newcite{collobert2011natural} used a convolutional architecture on a window surrounding the target token to be classified, considering several NLP tasks. 
%\kevincomment{targeted sentiment, others?}
%Our proposed encoders use similar architectural components.
%The main difference 
%between this prior work and ours 
%is that 
However, learning to represent tokens in supervised training can suffer from limited data. 
%they used supervised training for their token representation components and therefore had limited training data.  
%In contrast we are interested in using large unlabeled and multi-domain data sets and to learn more fine-grained context-sensitive representations. 
We instead focus on learning token embedding models on unlabeled data, then use 
%Our goal is to learn a token embedding model on unlabeled data, then use it 
them to produce features for downstream tasks. So we focus on efficient architectures and unsupervised learning criteria. 
%weakly-supervised and low-resource settings.  

The most closely related work consists of efforts to train LSTMs to represent tokens in context using unsupervised training objectives. 
%\ltcomment{discussion on ~\cite{peters2017semi} for learning sentence and document encoders from unlabeled data and ~\cite{melamud2016context2vec} learned a context encoder from unlabeled data for sentence completion, lexical substitution and word sense disambiguation}
\newcite{dyer-token-embed} use multilingual data to learn token embeddings that are predictive of their translation targets, while \newcite{melamud2016context2vec} and \newcite{peters2017semi} use unsupervised learning with monolingual sentences. 
%Their work is complementary to ours, showing how bilingual data and LSTM architectures can learn useful token representations for downstream tasks. We focus on the unsupervised setting in which we only require unannotated text in the language of interest, and we evaluate 
We experiment with LSTM token embedding models as well, though we focus on different tasks: POS tagging and dependency parsing. We generally found that very small contexts worked best for these syntactic tasks, thereby limiting the usefulness of LSTMs as token embedding models. 
%Most tasks pursued in this related work are 
%We will release our code and trained models to the community with the aim that they can be applied across tasks and domains. %\klcomment{include link here?}

\section{Token Embedding Models}
\label{sec:models}

%We begin with basic definitions. 
We assume access to pretrained type embeddings. 
Let $\vocab$ denote a vocabulary of word types. For each word type $x\in\vocab$, we denote its %$d$-dimensional 
type embedding by %(column) vector
$\vect{v}_x\in\mathbb{R}^{\typedim}$. 
% (KL) For all token embedding models, we 

We define a word sequence $\sent = \langle x_1, x_2,..., x_{|\sent|}\rangle$ in which each entry $x_j$ is a word type, i.e., $x_j\in\vocab$. 
%We assume access to a corpus of word sequences, each of which may be a sentence, paragraph, or document. We denote the corpus by $X = \{\senti \}_{i=1}^{|X|}$ where each $\senti$ is a word sequence. 
%We denote the $j$th element in sequence $i$ by $x_j^{(i)}$. 
We define a word token as an element in a word sequence. 
We consider the class of functions $\enc$ that take a word sequence $\sent$ and index $j$ of a particular token in $\sent$ 
and output a vector of dimensionality $\tedim$. 
We will refer to choices for $\encfull$ as \textbf{encoders}. 
%We denote the set of possible contexts of a word token by $\mathcal{C}$, so $f : \vocab \times \mathcal{C} \rightarrow \reals^{d'}$. 
%We now propose several choices for $\encfull$ that differ in terms of how context is handled. We will describe them as encoders since they encode a token and its context into a vector. This terminology also foreshadows the encoder-decoder style of training that we will describe below in Section~\ref{sec:losses}. 

%\begin{figure}[!t]
%\centering
%\begin{tabular}{cc}
%\hspace{0in}\includegraphics[width=2.3in]{fig/encoder_DNN} & \hspace{0.8in}\includegraphics[width=3in]{fig/encoder_CNN} \vspace{.1in}\\
%\vspace{.1in}
%\hspace{0in}\includegraphics[width=2.3in]{fig/encoder_biRtNN2} & \hspace{0in}\includegraphics[width=2.5in]{fig/encoder_DBM} \\
%\end{tabular}
% \vspace{-.2in}
%\caption{\small \label{fig:encoders} Encoder models.  (a) DNN encoder (b) word convolutional encoder (c) bidirectional recurrent encoder (d) deep Boltzmann machine.  For the examples here, all of the encoders except for the RNN use a window of 10 words of context to the left and right of the target word, while the RNN uses 3 words to the left and right of the target word.  The CNN here uses three filters with $n = 3$.  In all cases, the ellipsis stands for optional additional layers, and the light blue vector at the top is the token embedding for the target word.}
%\label{fig:encoders}
%\end{figure}

%\vspace{-0.2cm}
\subsection{Feedforward Encoders}
Our first encoder is a basic feedforward neural network 
%(Fig.~\ref{fig:encoders}a) 
that embeds the sequence of words contained in a window of text surrounding word $j$. We use a fixed-size window containing word $j$, the $\tewindow$ words to its left, and the $\tewindow$ words to its right. 
We concatenate the vectors for each word type in this window and apply an affine transformation followed by a nonlinearity:
%, i.e., $\vect{v}_{x_{j-k}},...,\vect{v}_{x_{j+k}}$, we stack them to obtain a $d(k+1)$-dimensional vector $\vect{c}$: 
%\begin{equation}
%\vect{c} = [\vect{v}_{x_{j-k}};...;\vect{v}_{x_{j+k}}]
%\end{equation}
%denote the columnwise concatenation of the column word vectors from position $j$ to position $\ell$ in $\sent$ by $[\vect{x}_j;...;\vect{x}_\ell]$. 
%We define the DNN encoder $\encseq$ as follows:
\begin{align}
& \encseqfull = \nonumber\\
& g\left(W^{(D)}[\vect{v}_{x_{j-\tewindow}};\vect{v}_{x_{(j-\tewindow)+1}};...;\vect{v}_{x_{j+\tewindow}}] + \vect{b}^{(D)}\right)\nonumber
\end{align}
\noindent where $g$ is an elementwise nonlinear function (e.g., $\tanh$), 
%$k$ is the number of context words to consider on either side, 
$W^{(D)}$ is a $\tedim$ by $\typedim(2\tewindow +1)$ parameter matrix, semicolon (;)
%$[\vect{a};\vect{b}]$ 
denotes vertical concatenation, 
%of vectors $\vect{a}$ and $\vect{b}$, 
and $\vect{b}^{(D)}\in\reals^{\tedim}$ is a bias vector. We assume that $\sent$ is padded with start-of-sequence and end-of-sequence symbols as needed. The resulting $\tedim$-dimensional token embedding can be transformed by additional nonlinear layers. %, making this a deep neural network (DNN) encoder.

This encoder does not distinguish word $j$ other than by centering the window at its position.  It is left to the training objectives to place emphasis on word $j$ as needed (see Section~\ref{sec:losses}). 
%For example, when minimizing reconstruction error in an autoencoding framework, we penalize the error of reconstructing word $j$ more than the other words. 
Varying $\tewindow$ will influence the phenomena captured by this encoder, with smaller windows capturing similarity in terms of local syntactic category (e.g., noun vs.~verb) and larger windows helping to distinguish word senses or to identify properties of the discourse (e.g., topic or style). 
%In our experiments below, we train DNN encoders and evaluate their embeddings qualitatively and quantitatively for part-of-speech tagging and dependency parsing. 

%An even simpler version of this encoder is a ``bag-of-words DNN'', obtained by adding the vectors of the $2\tewindow$ context words 
%instead of concatenating them,
%. This could be termed a ``bag-of-words'' encoder because it does not distinguish words in different positions; it only pays attention to whether a word is in the context of word $j$. This encoder 
%which may be better suited to large $\tewindow$ and broader discourse modeling, though likely weaker at capturing local syntactic/functional categories. 

%For a context window containing $k$ words on each side of the target word, this encoder performs the following to embed the $j$th token of sequence $\sent$:
%\begin{equation}
%f(\sent, j, k) = g\left(W^{(t)}\vect{v}_{x_j} + W^{(c)}\sum_{i=j-k, i\neq j}^{j+k}\vect{v}_{x_i} + \vect{b}\right)\nonumber
%\end{equation}
%\noindent 
%where $g$ is an elementwise nonlinear function such as $\tanh$, $W^{(t)}$ is a $d$-by-$d'$ parameter matrix that transforms the target word, $W^{(c)}$ is a $d$-by-$d'$ parameter matrix that transforms the sum of the context words, and $\vect{b}$ is a $d$-dimensional bias vector. We assume that $\sent$ is padded with start-of-sequence and end-of-sequence tokens as needed. The $d'$-dimensional token embedding $f(\sent, j, k)$ could be transformed by additional nonlinear layers. 

\subsection{Recurrent Neural Network Encoders}
%\ltcomment{I  added more description about seq2seq model}
The above feedforward DNN encoder will be brittle  
% (KL) when considering 
with large window sizes. 
We therefore also consider encoders based on recurrent neural networks (RNNs).  RNNs have recently enjoyed a great deal of interest in the deep learning, speech recognition, and NLP communities~\cite{sundermeyer2012lstm,graves2013speech,sutskever2014sequence}, 
%\cite{tai-socher-manning:2015:ACL-IJCNLP}, 
most frequently used with ``gated'' connections like long short-term memory (LSTM)~\cite{hochreiter1997long,gers2000learning}. 

We use an LSTM to encode the sequence of words containing the token and take the final hidden vector as the $\tedim$-dimensional encoding. While we can use longer sequences, such as the sentence containing the token~\cite{dyer-token-embed}, 
%.  However, for the time being 
we restrict the input sequence 
% (KL) to the LSTM to be 
to a fixed-size context window around word $j$, so the input is identical to that of the feedforward encoder above. For the syntactic tasks we consider, we did not find large context windows to be helpful. 

\subsection{Training}
\label{sec:losses}
 
We consider unsupervised ways to train the encoders described above. Throughout training for both models, the type embeddings are kept fixed. 
We assume that we are given a corpus $X = \{\senti \}_{i=1}^{|X|}$ of unannotated word sequences. 

One widely-used family of unsupervised criteria is that of reconstruction error and its variants. These are used when training autoencoders, which use an encoder $\enc$ to convert the input $\sent$ to a vector followed by a decoder $\dec$ that attempts to reconstruct the input from the vector.  
%Autoencoder-based training is most natural to use with our DNN encoder (Fig.~\ref{fig:encoders}a), but can also be used with other model architectures.
%Figure~\ref{fig:decoder} shows this architecture. 
The typical loss function is the squared difference between the input and reconstructed input.
%:
%. For our encoders, this takes the form: 
%\begin{equation}
%\lossre(\enc,\dec,\sent,j) = \sum_{i=1}^{|\sent|} \left\| \dec(\encfull)_i - \vect{v}_{x_i} \right\|_2^2 \nonumber
%\end{equation}
%We note that we write this loss differently from how it is usually written by explicitly summing over elements in the sequence $\sent$. We do so in order to 
We use a generalization that is sensitive to the position of elements.  Since our primary interest is in learning useful representations for a \emph{particular} token in its context, 
%while $\lossre$ weights the reconstruction of all tokens equally. So 
we use a weighted reconstruction error:
% (KL)
\vspace{-.08in}
\begin{equation}
%\textstyle
\losswre(\enc,\dec,\sent,j) = \sum_{i=1}^{|\sent|} \omega_i\left\| \dec(\encfull)_i - \vect{v}_{x_i} \right\|_2^2 \label{eq:reconst}
\end{equation}
%\vspace{-.1in}
\noindent where $\dec(\encfull)_i$ is the subvector of $\dec(\encfull)$ corresponding to reconstructing 
%the $i$th entry 
$\vect{v}_{x_i}$, and where $\omega_i$ is the weight for reconstructing the $i$th entry.
% in $\sent$. 
%This generalizes $\lossre$ by allowing the modeler to specify a fixed weighting vector $\vect{\omega}$ to target learning.  
%In pilot work (Sec.~\ref{sec:proposed_work_pilot}) we  found benefit from non-uniform weighting. 

For our feedforward encoder $\enc$, we use analogous fully-connected layers in the decoder $\dec$, forming a standard autoencoder architecture. 
%For our RNN encoder, we use an LSTM architecture for the decoder $\dec$ to reconstruct the input sequence of type embeddings. 
%with weighted squared loss. 
To train the LSTM encoder, we add an LSTM decoder to form a sequence-to-sequence (``seq2seq'') autoencoder~\cite{sutskever2014sequence,li-luong-jurafsky:2015:ACL-IJCNLP,dai-le-2015}. 
%, an LSTM-based learning architecture in both the inputs and outputs. 
That is, we use one LSTM as the encoder $\enc$ and another LSTM for the decoder $\dec$,  initializing $\dec$'s hidden state to the output of $\enc$. Since we use the same weighted reconstruction error described above, the decoder must output a single vector at each step rather than a distribution over word types. So we use an affine transformation on the LSTM decoder hidden vector at each step in order to generate the output vector for each step. Reconstruction error has efficiency advantages over log loss here in that it avoids the costly 
%Squared error has the advantage that it avoids a costly 
summation over the vocabulary. 
%the output type embedding, rather than using a softmax layer over output words. 

%While popular, reconstruction error requires a separate decoder that inverts the process of the encoder.
%For certain encoders, this is difficult or requires specialized decoder design~\cite{masci2011stacked,SocherEtAl2011:PoolRAE}. 
%Nonetheless, for most of our encoders, reconstruction error can be used or has already been adapted, so we plan to use it when possible. 
%Li et al.~\cite{li-luong-jurafsky:2015:ACL-IJCNLP} used conditional likelihood (also called cross-entropy/log loss) as the objective for reconstructing each word, and we plan to compare to it in future work. 
%Other losses are possible within the framework of reconstruction-based training, such as cross-entropy/log loss. 
%Squared error has the advantage that it avoids a costly summation over the vocabulary. 
%, and is still well-defined when the vocabulary is unbounded (e.g., when using subword information to define word type embeddings). 

%\input{losses}

\section{Qualitative Analysis}

Before discussing downstream tasks, we perform a qualitative analysis to show what our token embedding models learn.

\begin{table*}[t]
\centering
\small
\begin{tabular} {|c | l |c | l |}
\hline 
\qtok 
%@MENTION it's actually my 2nd ... 
my first \textbf{one was like \underline{2} minutes long and} has %%some
%issues , but as soon as i fix it up , i'll show it off ! 
& 
\qtok 
 \textbf{jus listenin \underline{2} mr hudson and} drake crazyness % !!!!!!!!
\\\hline
\ntok{1} 
%@MENTION italy is 
my fav \textbf{place- was there \underline{2} years ago and} am %%going
%back this year-wish i was there now
&
\ntok{1} 
@mention \textbf{deaddddd u go \underline{2} mlk high up} n %bk 
\\
\ntok{2} 
%it took james 6 months to get buried ??!! i 
thought it \textbf{was more like \underline{2} ..... either way} , i
%know his perm was still the bomb !!... lol .
& 
\ntok{2} 
%\#whyyoturkey breast 
only a \textbf{cups tho tryin \underline{2} feed the whole} family %\#nshit
\\
\ntok{3} 
%@MENTION nope been using it for awhile . closing an account for the first time . 
%%need 
to \textbf{backup everything from \underline{2} years before i} %%do %%that
%.\\
& 
\ntok{3}
%when i'm bored 
bored on mars \textbf{i kum down \underline{2} earth ... yupp} !!
\\
\ntok{4} 
i \textbf{slept for like \underline{2} sec lol .} freakin chessy
%woke me up . were having the munchies\\
& 
\ntok{4}
%rt @MENTION : every time when 
i miss \textbf{you i trying \underline{2} looking oud my} mind girl
%i miss you\\
\\
\hline
\hline
\qtok 
%new blog post : novel matters and between 
the lines \textbf{: i am \underline{so} thrilled about this} . may 
%i humbly ( or .. URL-tinyurl.com\\\hline
&
\qtok
fighting \textbf{off a headache \underline{so} i can work} on my 
%poster .... goal = finished tomorrow !\\\hline
\\\hline
\ntok{1}
%@MENTION no disruption at all . just chilling with my football 
and work \textbf{. i am \underline{so} glad you asked} . let 
%me drop you an email .\\
&
\ntok{1} 
im \textbf{on my phone \underline{so} i cant see} who @mention 
%is but it looks like omar\\
\\
\ntok{2} 
\textbf{i was \underline{so} excited to sleep} in tomorrow
% - not ! \#jewel\\
&
\ntok{2}
%worked out a little , not so fun . i 
did some \textbf{things that hurt \underline{so} i guess i} was doing 
%something right\\
\\
\ntok{3}
\textbf{@mention that is \underline{so} funny ! i} know which 
%song you are singing . cool as a tarot song . now i'll be singing it the rest of the day . lol\\
&
\ntok{3}
%rt @MENTION : @MENTION i'm sitting here chillin 
%%and 
my \textbf{phone keeps beeping \underline{so} i know ralph} must %%be 
%in the house showing luv .\\
\\
\ntok{4}
%one of my models has had the most beautiful 
little girl \textbf{! i was \underline{so} touched when she} called %%and 
%let me see the first photos ! i can't wait ...\\
&
\ntok{4}
%i just got 
randomly obsessed \textbf{with this song \underline{so} i bought it} %%lol %@MENTION 
% URL-t.co
\\
\hline
\end{tabular}
\caption{Query tokens of two polysemous words and their four nearest neighboring tokens. The target token is underlined and the encoder context (3 words to either side) is shown in bold. %For clarity, we show 
%Additional context words are shown as space permits. 
See text for details.}
\label{tab:qual_same_types}
\vspace{-0.3cm}
\end{table*}

\subsection{Experimental Setup}
\label{sec:qualsetup}

We train a feedforward DNN token embedding model on a corpus of 300,000 unlabeled English tweets. %\footnote{While we use a much larger dataset for training type embeddings, we found that we did not need to use a large training set for token embeddings, likely due to the parametric nature of our token embedding models.} 
%using a context window with $\tewindow=3$ words to the left and to the right of the token. 
We use a window size $\tewindow=3$ for the qualitative results reported here; for downstream tasks below, we will vary $\tewindow$. 
% (KL)
For training, we use our weighted reconstruction error (Eq.~\ref{eq:reconst}). 
The encoder uses one hidden layer of size 512 followed by the token embedding layer of size $\tedim = 256$. The decoder also uses a single hidden layer of size 512. We use ReLU activations %linear units 
%for the activations %function 
except the final encoder/decoder layers % and final encoder layer 
which use linear activations. 
%We use a corpus of 300,000 unlabeled English tweets as training data for the token embedding model. 

In preliminary experiments we compared 3 weighting schemes for $\vect{\omega}$ in the objective: for token index $j$,  ``uniform'' weighting sets $\omega_i=1$ for all $i$;  ``focused'' sets $\omega_j=2$ and $\omega_i=1$ for $i\neq j$; and ``tapered'' sets $\omega_j=4$, $\omega_{j\pm 1}=3$, $\omega_{j\pm 2}=2$, and 1 otherwise. The non-uniform schemes place more emphasis on reconstructing the target token,
% (KL) during learning, 
and we found them to slightly outperform uniform weighting. Unless reported otherwise, we use focused weighting for all experiments below.

We train using stochastic gradient descent with momentum for 1 epoch, saving the model that reaches the best objective value on a held-out validation set of 3,000 unlabeled tweets. 
For the type embeddings used as input to our token embedding model, we train 100-dimensional skip-gram embeddings on 56 million English tweets using the \texttt{word2vec} toolkit~\cite{NIPS2013_5021}.

\subsection{Nearest Neighbor Analysis}

%First, 
We inspect the ability of the encoder to distinguish different senses of ambiguous types.  
Table~\ref{tab:qual_same_types} shows query tokens (Q) followed by their four nearest neighbor tokens (with the same type), all from our held-out set of 3,000 tweets. 
%, all taken from a held-out set of 3,000 tweets. 
%, centered on the underlined word in boldface. 
We choose two polysemous words that are common in tweets: ``2'' and ``so''. As queries, we select tokens that express different senses. The word ``2'' %(upper half of the table) 
can be both a 
number (left) and a synonym of ``to'' (right).  The word ``so'' %(lower half)
is both an intensifier (left) and a connective (right). 
We find that the nearest neighbors, though generally differing in context words, have the same sense and same POS tag. 

In Table~\ref{tab:qual_diff_types} we consider nearest neighbors that may have different word types from the query type. For each query word, we permit the nearest neighbor search to consider tokens from the following set: \{``4'', ``for'', ``2'', ``to'', ``too'', ``1'', ``one''\}. In the first two queries, we find that tokens of ``4'' have nearest neighbors with different word types but the same syntactic category. That is, tokens of \emph{different} word types are more similar to the query than tokens of the \emph{same} type. We see this again with neighbors of ``2'' used as a synonym for ``to''. The encoder appears to be doing a kind of canonicalization of nonstandard word uses, which suggests applications for token embeddings in normalization of social media text~\cite{clark2011text}. See neighbor 8, in which ``too'' is understood as having the intended meaning despite its misleading surface form. 

\begin{table}[t]
\centering
\small
\begin{tabular}{|p{0.15cm}|l|}
\hline
\qtok %sunday , march 7th at 8:30 a.m. <URL-mastersswimmaui.org> the valley 
%isle 
masters \textbf{swimmers annual swim \underline{4} your heart !} %join us %!
\\
\hline
\ntok{1} 
%walked 
so \textbf{many miles loking \underline{for} her and handing} %out %fliers
\\
%, that i can barely walk , plus heel blisters . have her listed all over the place ,
\ntok{2} 
off to \textbf{the rehearsal space \underline{for} a weekend long} %studio %session 
\\
%. excited to work on some new material !
\ntok{3} %finally posted 
%my 
%thoughts 
\textbf{on the inauguration \underline{for} your enjoyment} % :} %URL-tinyurl.com 
\\
\hline
\qtok
\#canucks \textbf{now have a \underline{4} point lead on} the %red 
%wings after a win over the caps and a wings sol .
\\\hline
\ntok{1} %@mention idk my sis led he 
way lol \textbf{. it's the \underline{1} mile trail and} then you
%just walk across the river when you see it
\\
\ntok{2} %@mention it's actually my 2nd ... 
my first \textbf{one was like \underline{2} minutes long and} %has some 
%issues , but as soon as i fix it up , i'll show it off !
\\
\ntok{3} %@mention italy is 
my fav \textbf{place- was there \underline{2} years ago and} %am going 
%back this year-wish i was there now
\\
\hline
\qtok
\textbf{jus listenin \underline{2} mr hudson and} drake crazyness
%!!!!!!!!
\\
\hline
\ntok{1} 
@mention \textbf{deaddddd u go \underline{2} mlk high up} n bk \\
\ntok{2}  
%\#whyyoturkey breast 
only a \textbf{cups tho tryin \underline{2} feed the whole} family %\#nshit
\\
\ntok{3} 
\textbf{are ya'll listening \underline{to} the annointed one} ? he's on 
% tv again with the usual campaign lies
\\
\ntok{4}
@mention well \textbf{could u come \underline{to} mrs wilsons for} %a %lil bit ... im bringing some pizza rollz :) lol $<$3 janae
\\
\ntok{5} 
%when 
i'm bored 
on mars \textbf{i kum down \underline{2} earth ... yupp} !!
\\
\ntok{6}
\textbf{i am listening \underline{to} amar prtihibi -} black
\\
\ntok{7} 
%is 
%reading 
about \textbf{neopets and listening \underline{to} yelle ( URL} %URL-tinyurl.com} )
\\
\ntok{8} 
high ritee \textbf{now -\_\_\_\_\_\_- bout \underline{too} troop to the} crib %thoo
\\
\hline

	\end{tabular}
\caption{Nearest neighbors for token embeddings, where we consider neighbors that may have different word types from that in the query token. See text for details.}
	\label{tab:qual_diff_types}
\end{table}

\subsection{Visualization}

\begin{figure*}[t]
\includegraphics[width=1.0\textwidth]{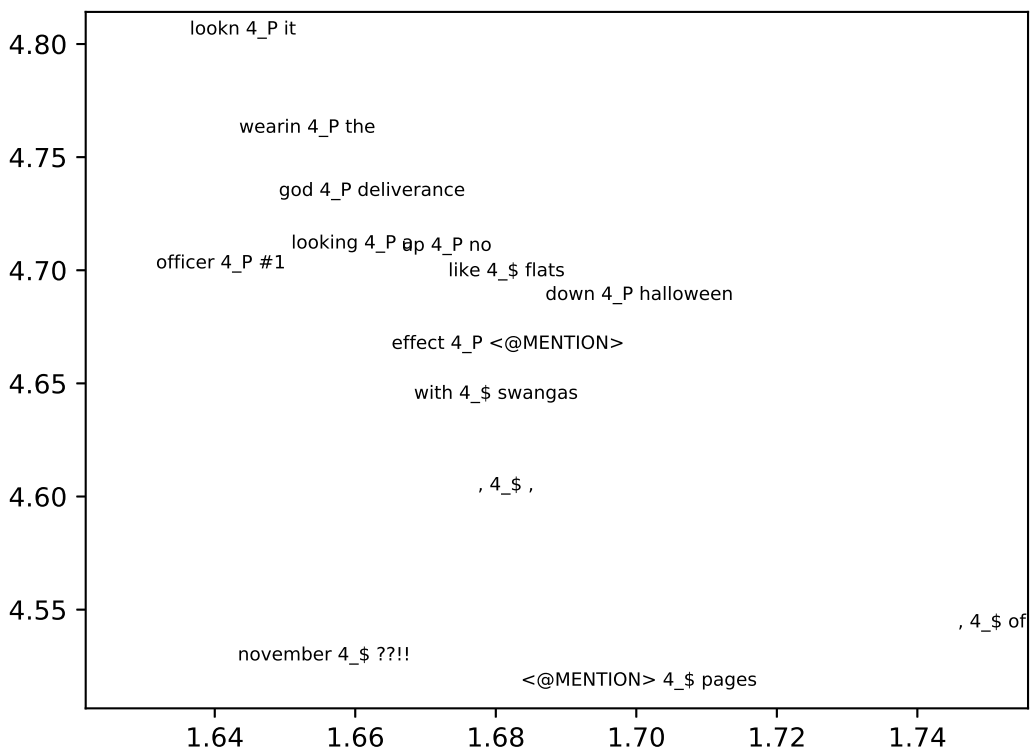}
\vspace{-1.2cm}
\caption{t-SNE visualization of token embeddings for word type ``4''. Each point shows the left and right context words ($\tewindow=1$) for the token along with the gold standard POS tag following an underscore (``\_''). The tag ``P'' is preposition and ``\$'' is number. Following the t-SNE projection, points were subsampled for this visualization for clarity.} 
%In this sample , and followed by the tag of the token in the tweet.} 
\label{fig:tsne}
\end{figure*}

In order to gain a better qualitative understanding of the token embeddings, we visualize the learned token embeddings using t-SNE~\cite{Hinton_visualizingdata}.  We learn token embeddings as above except with $\tewindow=1$. Figure~\ref{fig:tsne} shows a two-dimensional visualization of token embeddings for the word type ``4''. 
%Here we the get two-dimensional vectors from t-SNE, however we just plot the token embedding for the token $``4``$. 
%The window size is $\tewindow=1$ for the visualization of the learning the embedding.\footnote{Here we use $\tewindow=1$ because we get the best tagging performance in the following part-of-speech tagging task} Once finish training on the new setting, we can get the token embedding of every token in one sentence. 
For this visualization, we embed tokens in the POS-annotated tweet datasets from \newcite{gimpel-11a} and \newcite{owoputi-EtAl:2013:NAACL-HLT}, so we have their gold standard POS tags. 
We show the left and right context words (using $\tewindow=1$) along with the token and its gold standard POS tag. We find that tokens of ``4'' with the same gold POS tag are close in the embedded space, with prepositions appearing in the upper part of the plot and numbers appearing in the lower part. 
%And token of $``4``$ have different POS tag in different context can be easily separate. 

\section{Downstream Tasks}

%\vspace{-.05in}

We evaluate our token embedding models on two downstream tasks: POS tagging and dependency parsing. 
Given an input sequence $\sent = \langle x_1, x_2,..., x_{n}\rangle$, 
%$x=(x_{1},x_{2}, \dots, x_{n})$, 
we want to predict its tag sequence and dependency parse.  
We focus on Twitter since there is limited annotated data but abundant unlabeled data for training token embeddings.

\subsection{Part-of-Speech Tagging} 
\begin{figure}[t]
\small
%\label{tagger_model}
%\vspace{.3in}
%\centerline{\fbox{This figure intentionally left non-blank}}
\centering
\centerline{(a) Baseline DNN Tagger}
\includegraphics[width=0.4\textwidth]{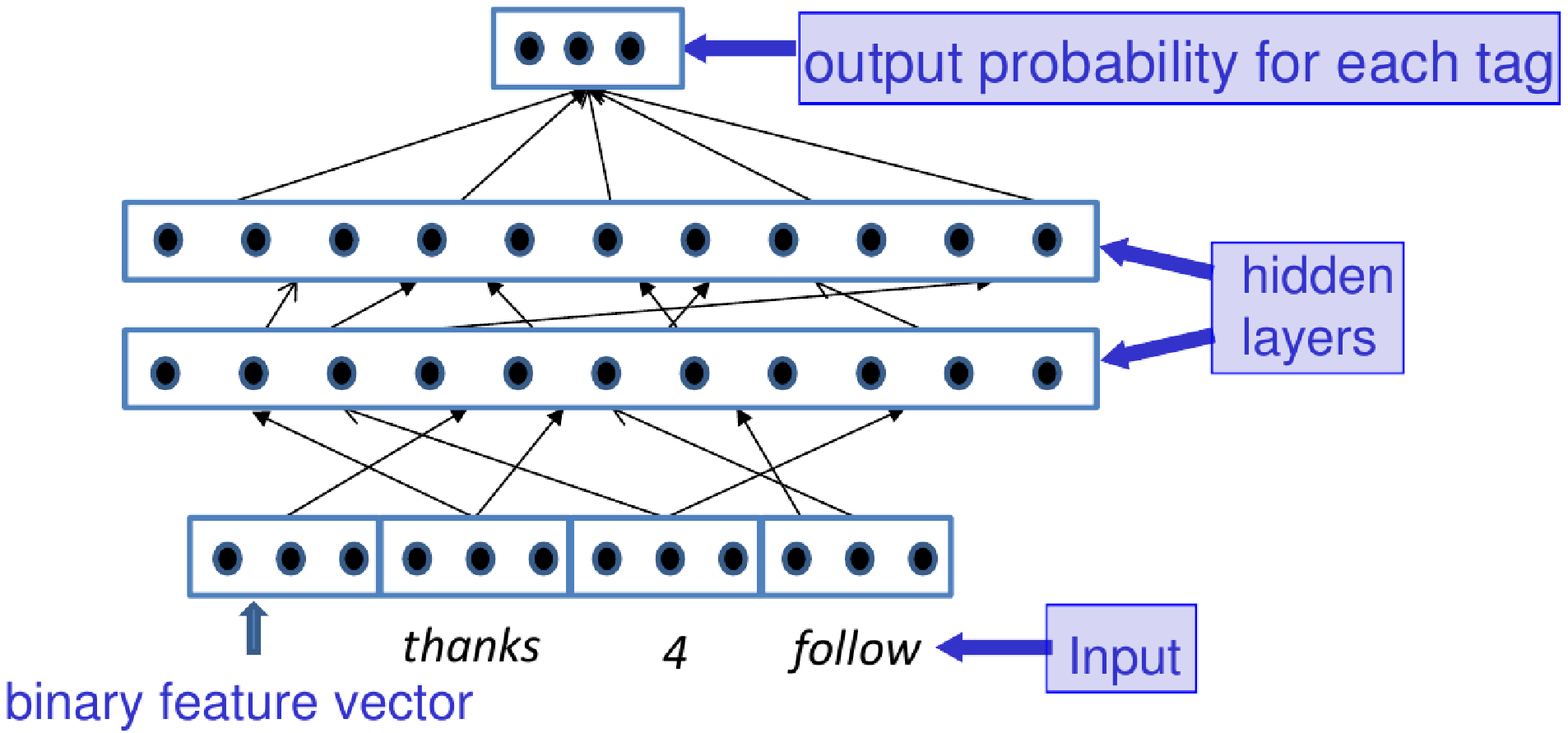}
\hspace{30ex}
\centerline{(b) Token Embedding Tagger}
\includegraphics[width=0.4\textwidth]{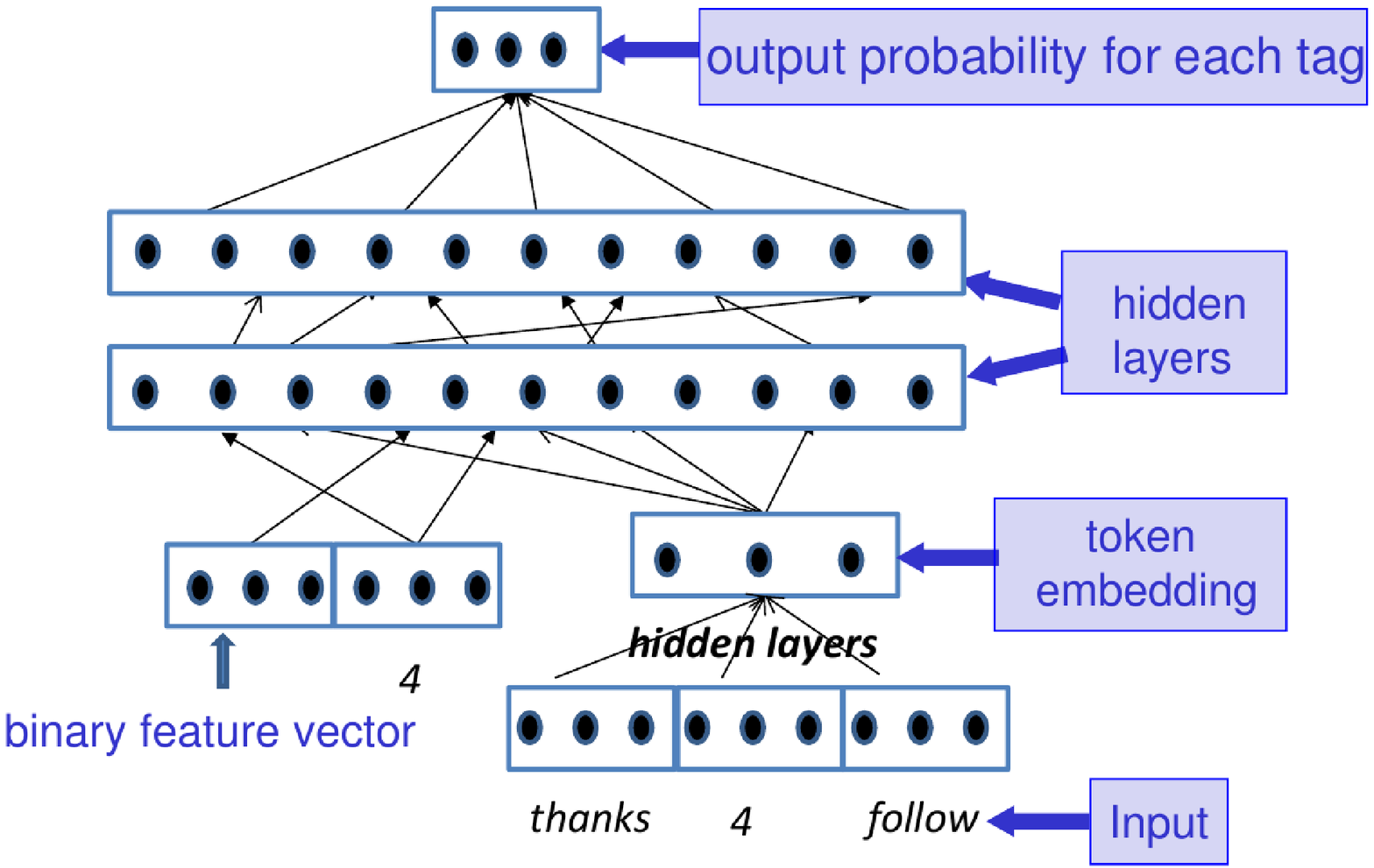}
%\includegraphics[width=0.4\textwidth]{./tagger_TE.png}
%\hspace{3ex}
%\vspace{.1in}
\caption{(a) Baseline DNN tagger, (b) tagger augmented with token embeddings.}
\vspace{-2ex} 
\label{tagger_model}
\end{figure}

\paragraph{Baseline}
We use a simple feedforward DNN as our baseline tagger. It is a local classifier that predicts the tag for a token independently of all other predictions for the tweet. That is, it does not use structured prediction. The input to the network is the type embedding of the word to be tagged concatenated with the type embeddings of $w$ words on either side.
%Below we experiment with different values of $w$. 
The DNN contains two hidden layers followed by one softmax layer. Figure~\ref{tagger_model}(a) shows this architecture for $w=1$ when predicting the tag of \emph{4} in the tweet \emph{thanks 4 follow}. %While not shown in the figure, w
We concatenate a 10-dimensional binary feature vector computed for the word being tagged (Table~\ref{table:word_feature}).\footnote{The definition of punctuation is taken from Python's \texttt{string.punctuation}.} 

\begin{table}[t]
\small
\centering
\begin{tabular}{|l|}%{|c|c|c|c|c|c|c|c|c|c|c|c|}
\hline
%@ & \# & rt & URL & digit & \$ & : & \dots & pun & puns \\
$x$ begins with @ and $|x|>1$ \\
%\hline
$x$ begins with \# and $|x|>1$ \\
%\hline
$\mathrm{lowercase}(x)$ is rt (retweet indicator)\\
%\hline
$x$ matches URL regular expression \\
%\hline
$x$ only contains digits \\
%\hline
$x$ contains \$ \\
%\hline
$x$ is : (colon)\\
%\hline
$x$ is \dots (ellipsis)\\
%\hline
%$x$ is punctuation and $|x|=1$ \\\hline
$x$ is punctuation and $|x|=1$ and x is not : or \$ \\
%\hline
$x$ is punctuation and $|x|>1$ and x is not \dots\\
\hline
\end{tabular}
\caption{Rules for %(10-dimensional) 
binary feature vector for word $x$. If multiple rules apply, the first has priority. The tagger uses this feature vector only for the word to be tagged; the parser uses one for the child 
and another for the parent in the dependency arc under consideration.}
\label{table:baseline_parser}
\label{table:word_feature}
\end{table}

We train the tagger by minimizing the log loss (cross entropy) on the training set, performing early stopping on the validation set, and reporting accuracy on the test set. We consider both learning the type embeddings (``updating'')  and keeping  them fixed. When we update the embeddings we include an $\ell_2$ regularization term penalizing the divergence from the initial type embeddings.

\paragraph{Token Embedding Tagger}
When using token embeddings, we concatenate the $\tedim$-dimensional token embedding to the tagger input.  
The rest of the architecture is the same as the baseline tagger. Figure~\ref{tagger_model}(b) shows the model when using type embedding window size $\blwindow=0$ and token embedding window size $\tewindow=1$. 

While training the DNN tagger with the token embeddings, we do not fine-tune the token embedding encoder parameters, leaving them fixed.

\subsection{Dependency Parser} 

\begin{table*}[t]
\centering
\begin{tabular}{|c|c|c|c|c|c|c|c|c|c|c|}
\hline
$\frac{i}{n}$ & $\frac{j}{n}$ & $\distance=1$ & $\distance=2$ & $3 \leq \distance \leq 5$ & $6 \leq \distance \leq 10$  & $\distance \geq 11$ & $i<j$ & $i>j$ & $x_j$ is wall symbol \\
\hline
\end{tabular}
\caption{Dependency pair features for arc with child $x_{i}$ and parent $x_{j}$ in an $n$-word sentence and where $\distance = |i-j|$. 
%The first two features are real-valued and the remaining are binary. %is the distance between the two words in the sentence. $i$ is the position for word $x_{i}$, $j$ is the position for word $x_{j}$, root = 1 means $x_{i}$ is the root in the sentence.
The final feature is 1 if $x_j$ is the wall symbol (\$), indicating a root attachment for $x_i$. In that case, all features are zero except for the first and last.
}
\label{table:parser_feature}
\end{table*}

\paragraph{Baseline} 
As our baseline, we use a simple DNN to do parent prediction independently for each word. That is, we use a local classifier that scores parents for a word. To infer a parse at test time, we independently choose the highest-scoring parent for each word. We also use our classifier's scores as additional features in TweeboParser~\cite{kong2014dependency}.

Our parent prediction DNN has two hidden layers and an output layer with 1 unit. This unit corresponds to a value $S(x_{i}, x_{j})$ 
that serves as the score for a dependency arc with child word $x_i$ and parent word $x_j$. The input to the DNN is the concatenation of the type embeddings for $x_{i}$ and $x_{j}$, the type embeddings of $w$ words on either side of $x_i$ and $x_j$, the features for $x_i$ and $x_j$ from Table~\ref{table:word_feature}, and features for the pair, including relative positions, direction, and distance (shown in Table~\ref{table:parser_feature}).\footnote{When considering the root attachment (i.e., $x_j$ is the wall symbol \$), the type embeddings for $x_j$ and its neighbors are all zeroes, the feature vector for $x_j$ is all zeroes, and the dependency pair features are all zeroes except the first and last.}

For a sentence of length $n$, the loss function we use for a single arc $(x_{i}, x_{j})$ follows:
\begin{align}
&\mathrm{loss}_{\mathrm{arc}}(x_{i}, x_{j})= \nonumber \\
& \;-S(x_{i},x_{j})+ \log\left(\sum_{k=0, k \neq i}^{n}\!\!\!\exp\{S(x_{i}, x_{k})\}\right)\label{eq:arcloss}
%
%, k \; is \; selected}
\end{align}
\noindent where $k=0$ indicates the root attachment for $x_i$. We sum over all possible parents even though the model only computes a score for a binary decision.\footnote{We found this to work better than only summing over the exponentiated scores of an arc or no arc for the pair $\langle x_i, x_j\rangle$.} Where $\mathrm{head}(x_i)$ returns the annotated parent for $x_i$, the loss for a sequence $\sent$ is:
\begin{align}
\sum_{i=1}^n 
\mathrm{loss}_{\mathrm{arc}}(x_{i}, \mathrm{head}(x_{i}))
%S(x_{i},\mathrm{head}(x_{i}))
\end{align}
\noindent After training, we predict the parent for a word $x_i$ 

as follows:
\begin{equation}
\overline{\mathrm{head}}(x_{i})= \argmax_{k\neq i}%, k \; is \; selected} 
S(x_{i},x_{k})
\label{eq:arcmax}
\end{equation}

%\vspace*{-2ex}
\paragraph{Token Embedding Parser}
For the token embedding parser, we use the $\tedim$-dimensional token embeddings for $x_i$ and $x_j$. We simply concatenate the two token embeddings to the input of the DNN parser. When $x_j = \$$, the token embedding for $x_j$ is all zeroes. 
%We also add the position of the words as the input the model. And for each word $x_{i}$ and $x_{j}$, we use one hot vector to represent the tag feature. And 
The other parts of the input are the same as the baseline parser. While training this parser, we do not optimize the token embedding encoder parameters.
%, leaving them fixed both during training and testing.
As with the tagger, we tune over the decision to keep type embeddings fixed or update them during learning, again using $\ell_2$ regularization when doing so. We tune this decision for both the baseline parser and the parser that uses token embeddings.

%\vspace{-.05in}
\section{Experimental Setup}

For training the token embedding models, we mostly use the same settings as in Section~\ref{sec:qualsetup} for the qualitative analysis. The only difference is that we train the token embedding models for 5 epochs, again saving the model that reaches the best objective value on a held-out set of 3,000 unlabeled tweets. 
We also experiment with several values for the context window size $\tewindow$ and the hidden layer size, reported below. % for the token embedding models. 
\subsection{Part-of-Speech Tagging}
%\vspace{-0.5ex}
We use the annotated tweet datasets from \newcite{gimpel-11a} and \newcite{owoputi-EtAl:2013:NAACL-HLT}. For training, we combine the 1000-tweet \octtrain set and the 327-tweet \octdev development set. For validation, we use the 500-tweet \octtest test set and for final testing we use the 547-tweet \dailytest test set. The DNN tagger uses two hidden layers of size 512 with ReLU nonlinearities %rectified linear units each
and a final softmax layer of size 25 %linear units 
(one for each tag). The input type embeddings are the same as in the token embedding model. We train using stochastic gradient descent with momentum and early stopping on the validation set. %(\octtest). 

\subsection{Dependency Parsing}
\vspace{-0.5ex}
We use data from \newcite{kong2014dependency}, dividing their 717 training tweets randomly into a 573-tweet train set and a 144-tweet validation set. 
%using their 573 tweets from the 717-tweet training set as our training data, a set of 144 tweets from the as our validation set for preliminary experiments and tuning, and their 
We use their 201-tweet \testnew as our test set. Kong et al.~annotated whether particular tokens are contained in the syntactic structure of each tweet (``token selection''). 
We use the same automatic token selection (TS) predictions as they did, which are 97.4\% accurate. 
%In the paper, it mentions that many elements in tweet have no syntactic function. These include hashtags, URLs, and emotion. 
%For the token selection part, we use their token selection program.
%Their first-order sequence model achieves 97.4\% token selection accuracy with either gold standard or automatic POS tags. 
We use a pipeline architecture in which unselected tokens are not considered as possible parents when performing the summation in Eq.~\ref{eq:arcloss} or the $\argmax$ in Eq.~\ref{eq:arcmax}. 

Like Kong et al., we use gold standard POS tags and gold standard TS during training and tuning. %We also use gold POS and gold TS when tuning. 
For final testing on \testnew, we use automatically-predicted POS tags and automatic TS (using their same automatic predictions %from Kong et al.~
for both). 
Like them, we use attachment $F_1$ score (\%) for evaluation. 
%And we use 100-dimensional zero vector as the embedding of root. And 256-dimensional zero vector as the token embedding of root. 
Our DNN parsers use two hidden layers of size 1024 with ReLU nonlinearities. 
%rectified linear units each. 
The final layer has size 1  
(the score $S(x_{i}, x_{j})$). We train using SGD with momentum.

%\subsection{Relation Extraction}
%We use 

\section{Results}
\subsection{Part-of-Speech Tagging}

\begin{figure}[t]
\vspace{-.2in}
%\centerline{\fbox{This figure intentionally left non-blank}}
\centering
\includegraphics[width=0.47\textwidth]{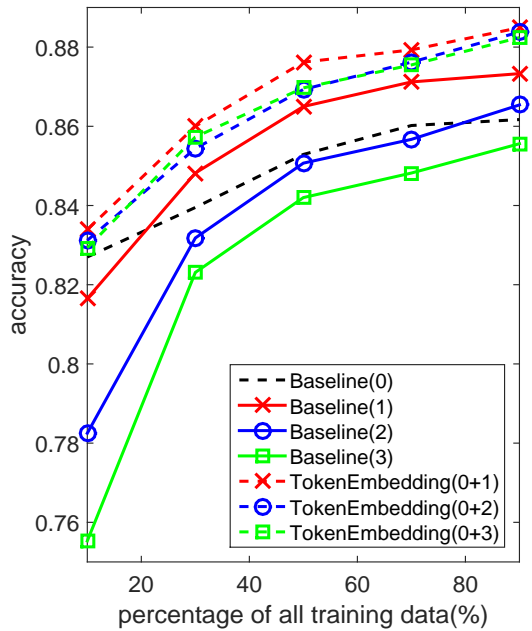}
%\hspace{3ex}
\vspace{-.05in}
\caption{
Tagging results.
``Baseline($w$)'' refers to the baseline tagger with context of $\pm w$ words; ``TokenEmbedding($w$+$w'$)'' refers to the token embedding tagger with tagger context of $\pm w$ words and token embedding context of $\pm w'$ words.}
%\klcomment{added description of notation}}
\label{fig:tagger_model_result}
\end{figure}

%\begin{table}[t]
%\centering
%\small
%\scalebox{0.8}{
%\begin{tabular}{c|c|c|}
%\cline{2-3}
%& OCT27TEST & DAILY547        \\ \hline
%\multicolumn{1}{|c|}{Baseline0} & 88.4  & 88.9            \\ 
%\hline
%\multicolumn{1}{|c|}{Baseline1} & 89.4  & 89.4  
%\\ \hline
%\multicolumn{1}{|c|}{Baseline2} & 89.16  & 89.34 
%\\ \hline
%\multicolumn{1}{|c|}{Baseline3} & 90.11  & 90.15 
%\\ \hline
%\multicolumn{1}{|c|}{Token Embedding Tagger0} & 89.98  & 89.76 
%\\ \hline
%\multicolumn{1}{|c|}{Token Embedding Tagger1} & 90.00  & 89.89
%\\ \hline
%\multicolumn{1}{|c|}{Token Embedding Tagger2} & 90.11  & 90.15 
%\\ \hline
%\multicolumn{1}{|c|}{Token Embedding Tagger3} & 90.44  & 90.52 
%\\ \hline
%\multicolumn{1}{|c|}{LSTM Token Embedding Tagger} & 90.57  & 90.68 
%\\ \hline
%\end{tabular}}
%
%\caption{Accuracies (\%) on development and test set using different models. Baseline1 : update the embedding during the training. Baseline2 : with one hot vector feature for each word. Baseline3 : update the embedding and use one hot vector feature. Token Embedding Tagger1: token embedding with updating the embedding. Token Embedding Tagger2: token embedding tagger with one hot vector feature.Token Embedding Tagger3: token embedding with updating the embedding and one hot vector feature . LSTM Token Embedding Tagger: updating the embedding and using one hot vector feature} 
%\label{table:final_tagger_result}
%\end{table}

We first train our baseline tagger without the binary feature vector 
%in Table~\ref{table:word_feature}) 
using different amounts of training data and window sizes $\blwindow \in \{0,1,2,3\}$. Figure~\ref{fig:tagger_model_result} shows accuracies on the validation set. 
%We compute the accuracy on the validation set for each fraction of training data and each $\blwindow$, showing the results in Figure~\ref{fig:tagger_model_result}. 
When using only 10\% of the training data, the baseline tagger with $\blwindow = 0$ performs best. As the amount of training data increases, the larger window sizes begin to outperform $\blwindow = 0$, and with the full training set, $\blwindow = 1$ performs best. 

Figure~\ref{fig:tagger_model_result} also shows the results of our token embedding tagger for $\blwindow=0$ and $\tewindow \in \{1,2,3\}$.\footnote{We used focused weighting for the results in Figure~\ref{fig:tagger_model_result} using $\omega_j=2$, but found slightly more stable results by increasing $\omega_j$ to 3, still keeping the other weights to 1. Our final tagging results use $\omega_j=3$.}  
We see consistent gains when using token embeddings, higher than the best baseline window for all values of $\tewindow$, though the best performance is obtained with $\tewindow =1$. When using small amounts of data, the baseline accuracy drops when increasing $\blwindow$, but the token embedding tagger is much more robust, always outperforming the $\blwindow = 0$ baseline.

We then perform experiments using the full training set, showing results in  Table~\ref{table:final_tagger_result}.
For all experiments with the baseline DNN tagger, we fix $\blwindow=1$; when using token embeddings, we fix $\blwindow=0$ and $\tewindow=1$. 
%We used these same $\blwindow$ and $\tewindow$ values for all other baseline and token embedding-augmented rows.
We also consider updating the initial word type embeddings during tagger training (``updating'') and using the binary feature vector for the center word (``features''). 
%We use new training set (OCT27TRAIN + OCT27DEV), new development set (OCT27TEST) and new test set(DAILY547). 
%We tuned $\blwindow$ and $\tewindow$ using the first setting shown in the table.  

Using token embeddings 
%We can see that the token embedding approach is 
consistently outperforms using type embeddings alone.  
%approach (e.g. standard word2vec embeddings) on this task. 
On the test set, we see gains from token embeddings across all settings, ranging from 0.5 to 1.2. The gains from DNN and seq2seq token embeddings are similar (possibly because we again use $\blwindow=0$ and $\tewindow=1$ for the latter). The baseline taggers improve substantially by updating type embeddings or adding features (settings (2) or (3)), but adding token embeddings still yields additional improvements. When we use token embeddings but remove the type embedding for the word being tagged (denoted ``*''), DNN TEs can still improve over the baseline, though seq2seq TEs yield lower accuracy. This suggests that the seq2seq TE model is focusing on other information in the window that is not necessarily related to the center word. 
%capturing something other than the 

\begin{table}[t]
\centering
\small
%\scalebox{0.95}{
%\begin{tabular}{|p{0.15cm}|l|c|c|}
\begin{tabular}{|l|c|c|}
\cline{2-3}
%\multicolumn{2}{l|}{} 
\multicolumn{1}{l|}{} & val. & test %\octtest & \dailytest        
\\ \hline
%1 & 
(1) Baseline & 88.4  & 88.9            \\ 
%2 & 
%$\:\:\:\:$
(1) + DNN TE & +1.6 & +0.9 \\ 
%89.98  & 89.76 \\ 
\hline
%2 & 
(2) Baseline + updating & 89.4  & 89.4  \\ 
%4 & 
(2) + DNN TE & +0.6 & +0.5 \\
%90.00  & 89.89\\ 
\hline
% 3 & 
(3) Baseline + features & 89.2  & 89.3 \\
% 6 & 
(3) + DNN TE* & +0.6 & +0.3 \\
(3) + DNN TE & +1.2 & +1.2 \\
\hline
% 3 & 
(3) Baseline + features & 89.2  & 89.3 \\
(3) + seq2seq TE* & -0.6 & -1.0 \\
(3) + seq2seq TE & +1.3 & +1.0 \\

%90.36  & 90.49 \\ 
\hline
% 7 & 
%(4) Baseline + updating + features  & 90.1  & 90.1 \\
%8 & 
%(4) + DNN TE  & +0.3 & +0.4 \\
%(4) + DNN TE & +0.33 & +0.37 \\%90.44  & 90.52 \\
%9 & 
%(4) + seq2seq TE + without updating & +0.4 & +0.2 \\\hline %90.57  & 90.68 \\ \hline
%10 & Owoputi et al., clusters + trans. & 89.50 & 90.54 \\
%& clusters + trans. + surface & 91.15 & 92.38 \\
%\hline
\end{tabular}
%}
\caption{Tagging accuracies (\%) on validation (\octtest) and test (\dailytest) sets. Accuracy deltas are always relative to the respective baseline in each section of the table. ``updating'' = updates type embeddings during training, ``features'' = uses binary feature vector for center word, * = omits center word type embedding. 
}
\label{table:final_tagger_result}
\end{table}

\begin{table}[t]
\centering
\small
%\scalebox{0.95}{
\begin{tabular}{|l|c|c|}
\cline{2-3}
%\begin{tabular}{|p{0.15cm}|l|c|c|}
%\cline{3-4}
\multicolumn{1}{l|}{} & val. & test %\octtest & \dailytest        
\\ \hline
%12 & 
(4) Baseline + all features & 92.1 & 92.2 \\
%& 91.76 & 91.39 \\
%13 & 12 
% 0.9212 0.9225
%0.9229 0.9252
(4) + updating & 92.2  & 92.4 \\ 
%& 92.02 & 92.02 \\
(4) + DNN TE + without updating & 92.4 & 92.8 \\
%(5) + LSTM token embeddings & 92.41 & 92.70 \\
%(5) + seq2seq TE & 0 & 0 \\

\hline
%11 & 
Owoputi et al. & 91.6 & 92.8 \\
\hline
\end{tabular}
%}
\caption{Tagging accuracies (\%) on validation (\octtest) and test (\dailytest) sets using all features: Brown clusters, tag dictionaries, name lists, and character $n$-grams. Last row is best result from Owoputi et al. (2013).}
\label{table:finalfinal_tagger_result}
\end{table}

\paragraph{Comparison to State of the Art.}
\newcite{owoputi-EtAl:2013:NAACL-HLT} achieve 92.8\%  on this train/test setup, using structured prediction and additional features from annotated and curated resources. 
We add several additional features inspired by theirs. We use features based on their generated Brown clusters, namely, binary vectors representing indicators for cluster string prefixes of length 2, 4, 6, and 8. 
We add tag dictionary features constructed from the Wall Street Journal portion of the Penn Treebank~\cite{marcus1993building}. We use the concatenation of the binary tag vectors for the three most common tags in the tag dictionary for the word being tagged. 
%If there are fewer than three tags given for the word, we pad with zeroes. 
We use the 10-dimensional binary feature vector and a binary feature indicating whether the word begins with a capital letter. All features above are used for the center word as well as one word to the left and one word to the right. 

We add several more features only for the word being tagged. We use name list features, adding a binary feature for each name list used by \newcite{owoputi-EtAl:2013:NAACL-HLT}, where the feature indicates membership on the corresponding name list of the word being tagged. 
We also include character $n$-gram count features for $n\in\{2,3\}$, adding features for the 3,133 bi/trigrams that appear 3 or more times in the tagging training data. 

After adding these features, we increase the hidden layer size to 2048. We use dropout, using a dropout rate of 0.2 for the input layer and 0.4 for the hidden layers. 
%No updating of word type embeddings is performed for these results. 
The other settings remain the same. 
The results are shown in Table~\ref{table:finalfinal_tagger_result}. Our new baseline tagger improves from 89.2\% to 92.1\% on validation, and improves further with updating. 

We then add DNN token embeddings to this new baseline. When doing so, we set $\blwindow=0$, as in all earlier experiments. 
%For each kind of token embedding, 
We add two sets of DNN token embedding features to the tagger, one with $\tewindow=1$ and another with $\tewindow=3$. 
%We use DNN token embeddings, 
%\footnote{Here we just report the result for DNN token embedding because they have similar performance from previous experiments.}, 
The results improve by 0.4 over the strongest baseline on the test set, 
%The resulting tagger achieves 92.76\% on the test set, improving over the baseline and 
matching the accuracy of \newcite{owoputi-EtAl:2013:NAACL-HLT}. This is notable since they used structured prediction while we use a simple local classifier, enabling fast and maximally-parallelizable test-time inference.  

\subsection{Dependency Parsing}

We show results with our head predictors in Table~\ref{table:window_parser_result}. 
The baseline head predictor actually does best with 
$\blwindow=0$. The predictors with token embeddings are able to leverage larger context: with DNN token embeddings, performance is best with $\tewindow=1$ while with seq2seq token embeddings, 
%\footnote{Because their performance are really close, we check their performance on test set, we find that the model with $\tewindow=1$ is better.}, 
performance is strong with $\tewindow=1$ and 2. 
When using token embeddings, we actually found it beneficial to drop the center word type embedding from the input, only using it indirectly through the token embedding functions. We use $\blwindow=-1$ to indicate this setting. 

The upper part of Table~\ref{table:final_parser_result_small} shows the results when we simply use our parsers to output the highest-scoring parents for each word in the test set. 
%For all experiments below we used $\blwindow=0$ for the baseline head predictor,  and $\blwindow=-1, \tewindow=1$. 
%We see that token embeddings improve performance. 
Token embeddings are more helpful for this task than type embeddings, improving performance from 73.0 to 75.8 for DNN token embeddings and improving to 75.0 for the seq2seq token embeddings.
  
We also use our head predictors to add a new feature to TweeboParser~\cite{kong2014dependency}. TweeboParser uses a feature on every candidate arc corresponding to the score under a first-order dependency model trained on the Penn Treebank. We add a similar feature corresponding to the arc score under our model from our head predictors. 
Because TweeboParser results are nondeterministic, presumably due to floating point precision, we 
%will be a little different for the trainings with the same hyper-parameters. With the new feature, We report means and standard deviations.
%We 
train TweeboParser 10 times for both its baseline configuration and all settings using our additional features, using TweeboParser's default hyperparameters each time. We report means and standard deviations.  

The final results are shown in the lower part of Table~\ref{table:final_parser_result_small}. 
%The fourth line shows the TweeboParser~\cite{kong2014dependency} performance is worse with less training data. 
%We also report the results when we add token embedding parser arc score features to TweeboParser and retrain it. %After adding the head predictor arc sore feature, it can gain better performance on validation set. 
While adding the feature from the baseline parser hurts performance slightly (80.6$\rightarrow$ 80.5), adding token embeddings improves performance. Using the feature from our DNN TE head predictor improves performance to 81.5, establishing a new state of the art for Twitter dependency parsing. 
%, while seq2seq token embeddings also improves performance to 81.0.

\begin{table}[t]
\centering
\small
\begin{tabular}{|c|c|c|c|}
\cline{2-4}
\hline
$\blwindow$ or $\tewindow$  & Baseline & DNN TE & seq2seq TE \\ \hline
\multicolumn{1}{|c|}{0} & \textbf{75.8} &  -    & -        \\ 
\hline
\multicolumn{1}{|c|}{1}  & 75.4  & \textbf{77.8} & 77.8
\\ %\hline
\multicolumn{1}{|c|}{2}  & 73.2 &   77.3 & \textbf{77.9}
\\ %\hline
\multicolumn{1}{|c|}{3} & 72.3  &  77.2 & 76.9 
\\ \hline

% 76.3, 76.9, 73.6, 72.7
\end{tabular}
\caption{Attachment $F_1$ (\%) on validation set 
%(sampled 144 tweets from 717 tweets with Gold POS and TS) 
using different models and window sizes. %The column "baseline1" shows the result for the baseline model, however training on whole 717 tweets and validation on \testnew . 
For TE columns, the input does not include any type embeddings at all, only token embeddings. 
Best result in each column is in boldface.} 
\label{table:window_parser_result}
\end{table}

%\begin{table}[t]
%\centering
%\small
%\begin{tabular}{|l|c|c|}
%\cline{2-3}
%\multicolumn{1}{c|}{} & val. & test \\
%\hline
%(1) Baseline parser & 75.1 & 74.2 \\
%(1) Baseline parser($\blwindow=0$) & 75.8 & 73.0 \\
%(1) + DNN TE($\tewindow=1$) & 77.8 & 75.8 \\
%(1) + seq2seq TE($\tewindow$=2) & 77.9 & 74.2 \\
%(1) + seq2seq TE($\tewindow$=1) & 77.8 & 75.0 \\
%\hline
%(2) Kong et al.  & - & 80.6 $\pm$ 0.13  \\
%(2) Kong et al.(less data)  & 78.0 $\pm$0.31  & 80.6 $\pm$ 0.25  \\
%(2) + Baseline parser($\blwindow=0$) & 78.0 $\pm$ 0.44 & 80.5 $\pm$ 0.30\\
%(2) + DNN TE($\tewindow=1$) & 78.1 $\pm$ 0.31 & 81.5 $\pm$ 0.25 \\
%(2) + seq2seq TE($\tewindow=2$) & 78.0 $\pm$ 0.25 & 80.9 $\pm$ 0.33  \\
%(2) + seq2seq TE($\tewindow=1$) & 77.9 $\pm$ 0.31 & 81.0 $\pm$ 0.17  \\
%\hline
%\end{tabular}
%\caption{Dependency parsing unlabeled attachment $F_1$ (\%) on validation and test (\testnew) sets for baseline parser and token embedding parsers.
%}
%\label{table:final_parser_result_small}
%\end{table}

\begin{table}[t]
\centering
\small
\begin{tabular}{|l|c|}
\hline
(1) Baseline parser ($\blwindow=0$)  & 73.0 \\
(1) + DNN TE ($\blwindow=-1, \tewindow=1$)  & 75.8 \\
(1) + seq2seq TE ($\blwindow=-1, \tewindow=1$) & 75.0 \\
(1) + seq2seq TE ($\blwindow=-1, \tewindow=2$) &  74.2\\
\hline
%(2) Kong et al.   & 80.6 $\pm$ 0.13  \\
%(2) Kong et al. (less data)    & 80.6 $\pm$ 0.25  \\
(2) Kong et al. & 80.6 $\pm$ 0.25  \\
(2) + Baseline parser ($\blwindow=0$)  & 80.5 $\pm$ 0.30\\
(2) + DNN TE ($\blwindow=-1, \tewindow=1$) & 81.5 $\pm$ 0.25 \\
%(2) + seq2seq TE($\tewindow=2$) & 78.0 $\pm$ 0.25 & 80.9 $\pm$ 0.33  \\
(2) + seq2seq TE ($\blwindow=-1, \tewindow=1$) & 81.0 $\pm$ 0.17  \\
(2) + seq2seq TE ($\blwindow=-1, \tewindow=2$) & 80.9 $\pm$ 0.33 \\
\hline
\end{tabular}
\caption{Dependency parsing unlabeled attachment $F_1$ (\%) on test (\testnew) sets for baseline parser and results when augmented with token embedding features. Following Kong et al., we report three significant digits.
}
\label{table:final_parser_result_small}
\end{table}

\section{Conclusion}
We have presented a simple and efficient way of learning representations of words in their contexts using unlabeled data, and have shown how they can be used to improve syntactic analysis of Twitter. Qualitatively, our token embeddings are shown to encode sense and POS information, grouping together tokens of different types with similar in-context meanings.  
Quantitatively, using token embeddings in simple predictors consistently improves performance, even rivaling the performance of strong structured prediction baselines.  
Our code and trained token embedding models are publicly available at the authors' websites. 
Future work includes further exploration of token embedding models, unsupervised objectives, and their integration with supervised predictors. 

\section*{Acknowledgments}
% Add any acknowledgments below. We'll uncomment this section in the final version only.
We thank the anonymous reviewers, Chris Dyer, and Lingpeng Kong. 
We also thank the developers of Theano~\cite{2016arXiv160502688short} and  Lasagne~\cite{lasagne} as well as NVIDIA Corporation for donating GPUs used in this research. 

% include your own bib file like this:
%\bibliographystyle{acl}
%\bibliography{acl2017}
\bibliography{acl2017}
\bibliographystyle{acl_natbib}

\appendix

\end{document}